\documentclass[10pt,twocolumn,letterpaper]{article}

\usepackage[pagenumbers]{iccv} 
\usepackage{bbding,pifont}
\definecolor{iccvblue}{rgb}{0.21,0.49,0.74}
\usepackage[pagebackref,breaklinks,colorlinks,allcolors=iccvblue]{hyperref}

\def\paperID{11301} 
\def\confName{ICCV}
\def\confYear{2025}

\usepackage{multirow}
\usepackage{float}

\title{RefSTAR: Blind Facial Image Restoration with\\Reference Selection, Transfer, and Reconstruction}

\author{
Zhicun Yin$^{1}$ \quad Junjie Chen$^{1}$ \quad Ming Liu$^{1(}$\Envelope$^)$ \quad Zhixin Wang$^{2}$ \\ \quad Fan Li$^{2}$ \quad Renjing Pei$^{2}$ \quad Xiaoming Li$^{3}$ \quad Rynson W.H. Lau$^{4}$ \quad Wangmeng Zuo$^{1}$ \\
$^{1}$ Harbin Institute of Technology \quad $^{2}$ Huawei Noah’s Ark Lab \\ \quad $^{3}$ Nanyang Technological University
\quad $^{4}$ City University of Hong Kong\\
{\tt\small \href{mailto:cszcyin@outlook.com}{\color{black}cszcyin@outlook.com}, \href{mailto:csmliu@outlook.com}{\color{black}csmliu@outlook.com}, \href{mailto:csxmli@gmail.com}{\color{black}csxmli@gmail.com}, \href{mailto:wmzuo@hit.edu.cn}{\color{black}wmzuo@hit.edu.cn}}
}

\begin{document}

\maketitle

\maketitle

\begin{abstract}

Blind facial image restoration is highly challenging due to unknown complex degradations and the sensitivity of humans to faces.
Although existing methods introduce auxiliary information from generative priors or high-quality reference images, they still struggle with identity preservation problems, mainly due to improper feature introduction on detailed textures.
In this paper, we focus on effectively incorporating appropriate features from high-quality reference images, presenting a novel blind facial image restoration method that considers \textbf{ref}erence \textbf{s}election, \textbf{t}ransfer, \textbf{a}nd \textbf{r}econstruction (RefSTAR).
In terms of selection, we construct a reference selection (RefSel) module.
For training the RefSel module, we construct a RefSel-HQ dataset through a mask generation pipeline, which contains annotating masks for 10,000 ground truth-reference pairs.
As for the transfer, due to the trivial solution in vanilla cross-attention operations, a feature fusion paradigm is designed to force the features from the reference to be integrated.
Finally, we propose a reference image reconstruction mechanism that further ensures the presence of reference image features in the output image.
The cycle consistency loss is also redesigned in conjunction with the mask.
Extensive experiments on various backbone models demonstrate superior performance, showing better identity preservation ability and reference feature transfer quality.
Source code, dataset, and pre-trained models are available at \url{https://github.com/yinzhicun/RefSTAR}.

\end{abstract}
\begin{figure*}[h]
	\centering
        \footnotesize
        \setlength\abovecaptionskip{0.2\baselineskip}
        \setlength\belowcaptionskip{0.2\baselineskip}
	\includegraphics[width=1.0\linewidth]{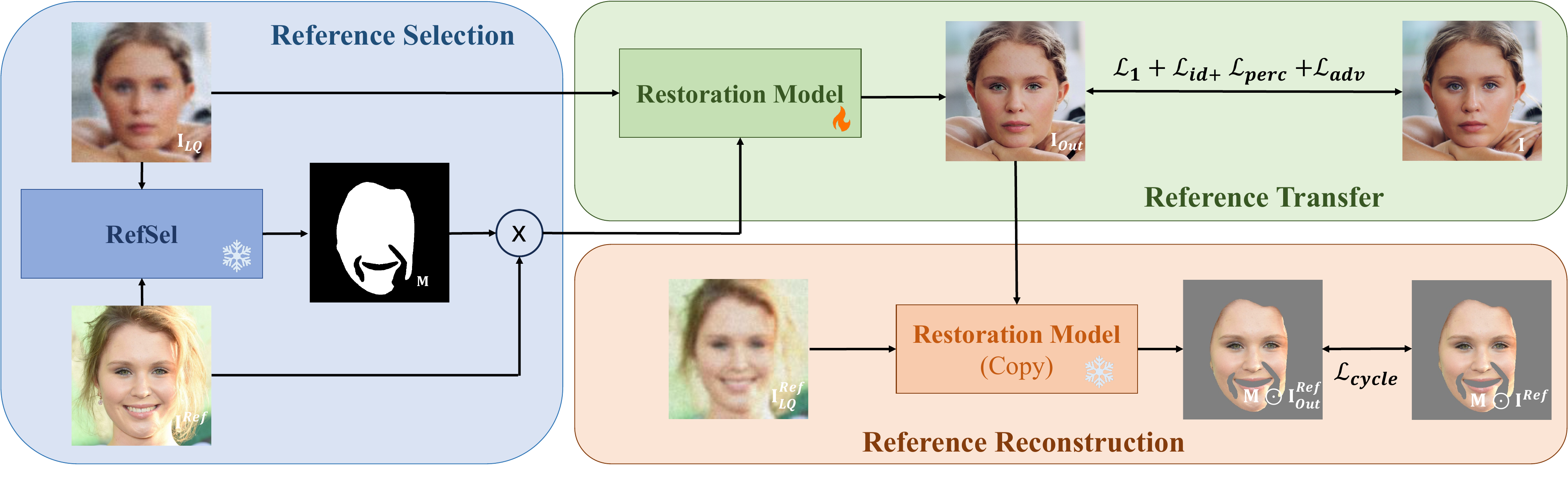}
    \caption{Pipeline of our proposed RefSTAR framework, including reference selection, transfer, and reconstruction. With the RefSel selecting proper reference features, the restoration model conducts the reference feature alignment under the supervision of our Mask-Compatible Cycle-Consistency Loss for reference reconstruction. All components make sufficient utilization of proper reference features.}
    \label{fig:RefSTAR_pipeline}
    
\end{figure*}
\section{Introduction}
\label{sec:intro}

Blind face image restoration aims to recover high-quality facial images from degraded inputs, without knowing the degradation type and parameters. While current methods~\cite{Li_2020_ECCV,GPEN,GFPGAN,CodeFormer,wang2022restoreformer,DifFace,DiffBIR,OSDFace} have made significant strides in generating plausible facial structures, they still face challenges in preserving identity and faithfully restoring personalized textures, 
particularly when dealing with severely degraded inputs. 
As a result, reference-based face restoration methods~\cite{GFRNet,ASFFNet,DMDNet,RefLDM,RestoreID,FaceMe,InstantRestore} have emerged as a promising solution, using high-quality images of the same person to provide identity-related features.

Ideally, when the reference and ground-truth images have perfectly identical textures, these methods can effectively transfer textures from the reference. However, during actual training, images of the same person often exhibit various differences, such as poses, facial expressions, and textures corresponding to different age stages. These discrepancies make the network uncertain about whether certain textures should be transferred. 
As a result, these methods struggle to effectively utilize reference images for more faithful restoration.
An alternative approach is to constrain the model using one of the GT and the reference image that more closely resembles the output. However, the optimization trajectory tends to lean heavily towards the reference image, as the model struggles to recover details similar to the GT, especially when the input is severely degraded.

Therefore, how to effectively leverage reference images for faithful restoration is an important yet underexplored issue. To address this challenge, we propose a new framework, \ie, RefSTAR, which methodically addresses reference texture selection, transfer, and reconstruction to improve restoration fidelity.
Specifically, to achieve effective texture selection, we design a RefSel module to explicitly predict which regions of the reference image should be selected for texture transfer. For instance, if the reference image shows an open mouth and the input LR image features a closed mouth, the model needs to exclude the teeth from the reference image. 
To train RefSel, we manually annotated 10,000 pairs of ground truth and reference images, marking the regions where the textures are consistent between them. By randomly degrading the ground-truth images, we use the low-quality images along with the reference images as input of RefSel to predict the texture consistency regions. 
Notably, when the input image is severely degraded, we use a full-face mask as the ground-truth region, as the visible textures in such degraded input cannot be reliably matched with the reference image, making it challenging to identify consistent texture regions.
In contrast to other methods, which implicitly learn texture selection through optimization between the restored image and the ground truth, our explicit approach is more effective in 
ensuring accurate restoration by directly identifying and transferring relevant features from the reference region.

After identifying consistent texture regions, the next challenge is how to align the selected reference textures with the degraded images to facilitate effective feature transfer. In the diffusion-based models~\cite{LDM,ReferenceNet}, a straightforward and widely adopted approach is to use cross-attention mechanisms. However, standard cross-attention struggles with misalignment and can lead to suboptimal feature fusion, where the model either ignores useful reference details or naively copies textures without proper adaptation. To address this issue, we use a dual-stream attention mechanism~\cite{IP-Adapter} that preserves feature hierarchy independence while enabling proper cross-stream interaction. By carefully guiding the fusion process, our approach prevents the model from indiscriminately copying reference features and instead facilitates a more context-aware and structurally coherent texture transfer, leading to higher-fidelity restoration.

Finally, to further enhance the transfer of reference textures, we introduce a cycle consistency constraint that aids in reconstructing the reference image from the restored results.
Specifically, the restored image is used as a new reference, while the original reference is degraded to serve as the low-quality input.
This cycle consistency encourages the model to effectively preserve and transfer relevant textures from the reference image, thereby improving the faithful restoration of the personalized details.

Experimental results demonstrate the effectiveness of our RefSel module in identifying texture-consistent regions, which in turn facilitates more precise and fine-grained texture transfer. 
As a result, our method achieves promising performance on both synthetic and real-world datasets, showcasing its robustness in handling diverse degradation scenarios and improving the fidelity of restored images.
The main contributions are summarized as follows.
\begin{itemize}
    \item We propose RefSTAR, a new framework for reference-guided face restoration, which explicitly addresses inconsistencies between degraded input and reference images. To support this, we manually annotate consistency regions in RefSel-HQ, a dataset of 10,000 image pairs.
    \item We introduce a new  reconstruction process, enhanced by cycle consistency, to ensure the effective transfer of relevant reference textures.
    \item Extensive experiments are conducted on both synthetic and real-world images, demonstrating the effectiveness of the presented RefSTAR method in terms of identity preservation ability and reference feature transfer quality.
\end{itemize}
\section{Related Works}
\subsection{Blind Face Image Restoration}

Blind face image restoration is particularly challenging due to unknown degradation and the difficulty of simulating real-world degradations~\cite{sharipov2023blind,li2023survey}. Therefore, existing methods leverage structured priors to guide the restoration process.
Early approaches primarily rely on geometric priors, such as face landmarks~\cite{Super-fan, Fsrnet, PFSR}, parsing maps~\cite{SAST, DSFD, Hifacegan}, and component heatmaps~\cite{FSRG}, to enforce semantic consistency in the reconstructed face structures. However, these methods struggle to generalize under complex real-world degradations.
With the development of generative models, generative priors have been widely adopted, particularly those from GANs~\cite{StyleGAN, StyleGAN2, VQGAN} and diffusion models~\cite{LDM}, which capture richer structural and textural details.
GAN-based approaches exploit latent space representations to synthesize photorealistic textures. For instance, GFP-GAN~\cite{GFPGAN} and GPEN~\cite{GPEN} leverage the latent space of StyleGAN~\cite{StyleGAN} to restore facial details, while CodeFormer~\cite{CodeFormer} utilizes a codebook-based generative prior~\cite{VQGAN} to achieve a balance between fidelity and realism.
Diffusion-based methods take a different approach by iteratively refining the restoration results through a denoising process. Models such as DifFace~\cite{DifFace}, DiffBIR~\cite{DiffBIR}, and PGDiff~\cite{PGDiff} demonstrate great restoration performance by progressively reconstructing missing details. To improve inference efficiency, OSDFace~\cite{OSDFace} employs a one-step diffusion strategy with a visual representation embedder, significantly reducing computational overhead.
While these methods achieve impressive results with powerful generative priors, their fidelity remains fundamentally constrained by the amount of useful information present in the degraded input.

\subsection{Reference-based Face Restoration}
For faithful face restoration, Li~\etal~\cite{GFRNet, ASFFNet} first introduce the use of high-quality images of the same identity to guide identity recovery and preserve personalized details. Early reference-based methods enhanced restoration by directly incorporating residual features~\cite{GFRNet, ASFFNet} or constructing reference dictionaries~\cite{DMDNet}. While these approaches outperform methods without reference images, they struggle to preserve fine-grained textures due to their limited ability to selectively transfer reliable details from the reference images.
To address this limitation, recent methods incorporate personalized priors from diffusion-based customized image generation models~\cite{DreamBooth,Celeb-Basis,stableidentity}. By integrating reference features into the iterative denoising process~\cite{RefLDM, RestoreID, MGFR, FaceMe}, these models effectively exploit identity information while benefiting from generative priors to handle extreme degradation. Moreover, InstantRestore~\cite{InstantRestore} employs a one-step inference strategy, achieving near real-time restoration with strong identity preservation.
Despite these advancements, most reference-based methods rely heavily on direct ground-truth supervision. Variations in lighting, pose, and facial expressions often introduce inconsistencies between the ground truth and reference images, causing models to disregard valuable reference details. Consequently, these approaches struggle to synthesize high-fidelity, personalized textures.
{Recently, RefineIR~\cite{RefineIR} proposed a loss formulation, which constrains the model with either the ground truth or the reference, based on which one is more similar to the output. However, recovering the details resembling the ground truth is difficult, especially when the input quality is poor, making the optimization bias towards the reference and resulting in excessive dependency that limits the generalization ability.}
Additionally, reliance on face landmark detection and image warping imposes further constraints, reducing robustness in cases where landmarks cannot be accurately extracted.
In summary, identifying which regions of the reference image to leverage and how much texture to transfer remains a fundamental challenge.
In this work, we introduce a new framework that explicitly selects and transfers reference features, enabling effective guidance while ensuring robust reconstruction quality.

\section{Methods}
\label{sec:method}

As shown in \cref{fig:RefSTAR_pipeline}, to incorporate appropriate features from high-quality reference images, our presented \textbf{RefSTAR} considers reference-guided face restoration from three aspects, namely \textbf{ref}erence \textbf{s}election, \textbf{t}ransfer, \textbf{a}nd \textbf{r}econstruction.
The reference selection (RefSel) module predicts a binary mask to point out the regions whose features can be migrated to the final output, mainly based on the conflict level between the input and the reference image.
The transfer component provides a pathway for features from the reference image to be infused into the restoration model, where a dual-stream cross-attention mechanism is deployed to force the infusion while avoiding explicit image/feature alignment.
Finally, the reconstruction process redesigns the cycle consistency loss as Mask-Compatible Cycle-Consistency Loss, which compels the model output to contain detailed textures from the reference and is consistent with the predicted binary mask.
More details about our RefSTAR are given in the following.

\subsection{Reference Feature Selection}
\label{sec:RefSTAT_method_RefSel}

To ensure an effective reference selection in our RefSel module, we first construct a data engine to obtain the input, reference, and binary mask triplets, \ie, $\{\mathbf{I}_\mathit{LQ}, \mathbf{I}^\mathit{Ref}, \mathbf{M}\}$, based on which RefSel can be effectively optimized.

\begin{figure}[t]
	\centering
        \footnotesize
       
	\includegraphics[width=.9\linewidth]{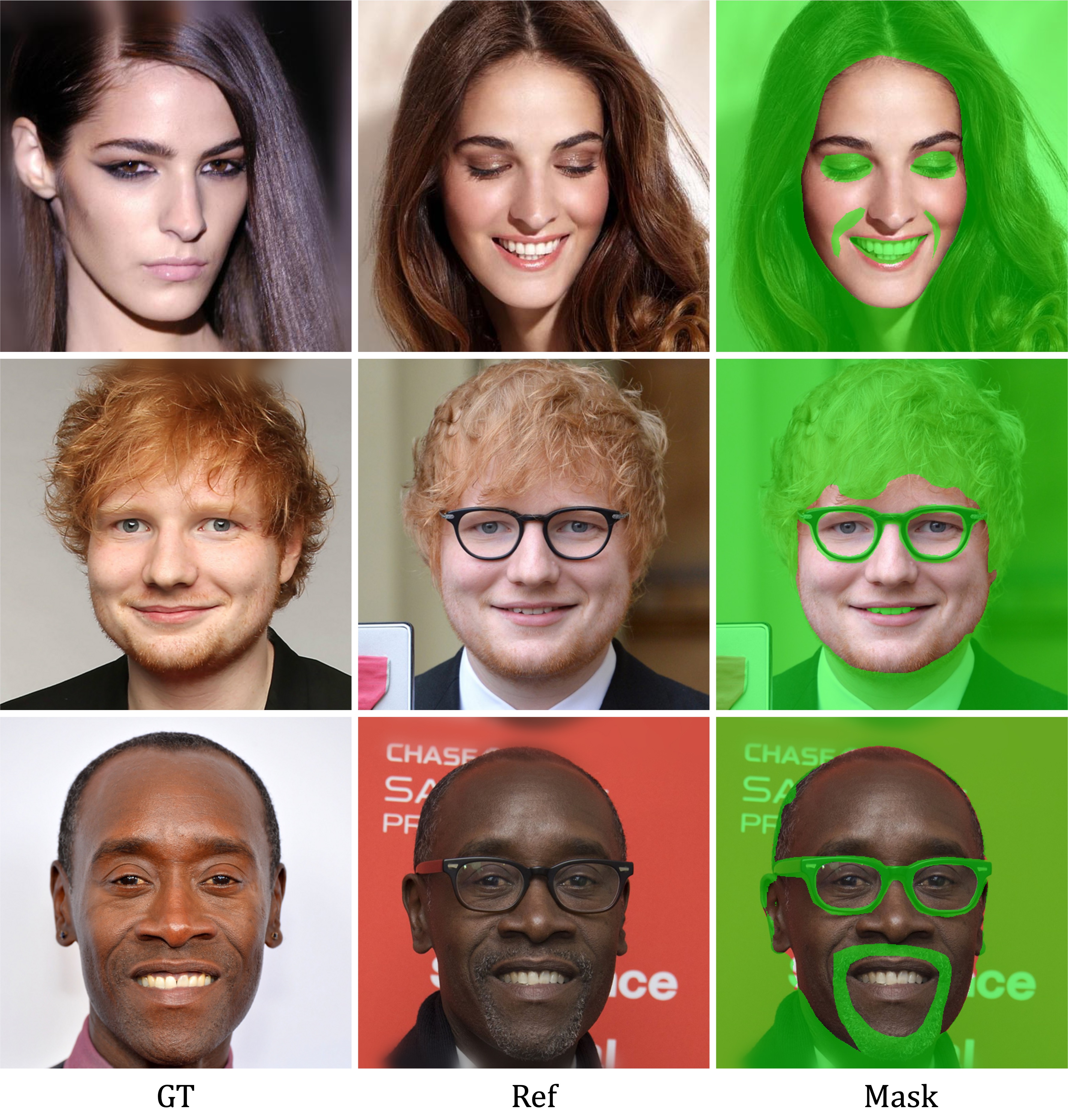}
    \caption{Examples of Annotation Pairs in RefSel-HQ Dataset.}
    \label{fig:RefSTAR_gt-ref-mask}
\end{figure}

\noindent \textbf{Data Engine.}
To begin with, we collect over 10,000 high-quality image pairs from existing facial image datasets (\eg, CelebRef-HQ~\cite{DMDNet}), serving as the ground truth $\mathbf{I}$ and reference $\mathbf{I}^\mathit{Ref}$, respectively.
For effectively obtaining the mask $\mathbf{M}$ for sufficient $\{\mathbf{I},\mathbf{I}^\mathit{Ref}\}$ pairs, a small portion (\ie, 800) of the image pairs are manually annotated, based on which a binary segmentation U-Net~\cite{U-Net} is trained.
Subsequently, we generate masks using the U-Net and concurrently perform a manual filtering and adjusting process until we obtain 10,000 high-quality $\{\mathbf{I},\mathbf{I}^\mathit{Ref},\mathbf{M}\}$ triplets (at this point, we have generated approximately 12,000 masks).
For generating low-quality input $\mathbf{I}_\mathit{LR}$, we adopt the degradation model of Real-ESRGAN~\cite{Real-ESRGAN}, with motion and defocus deblurring embedded for a more realistic degradation.
The mask annotation protocol adhered to a dual-axis taxonomy, \ie, the regions with the following conflicts are discarded.
\begin{itemize}
\item Dynamic conflicts capturing expression-induced mismatches (\eg, mouth open/close, eye open/close, and perioral wrinkle discrepancies during the variation of different expressions like smile and frown).
\item Static conflicts with feature misalignments arising from physiological traits (\eg, pigmentation disparities like freckles) or some artificial accessories (\eg, eyewear occlusion and texture clashes from heavy cosmetics).
\end{itemize}
Some examples of our data set are shown in \cref{fig:RefSTAR_gt-ref-mask}. Note that all the backgrounds and hair are masked to reduce the complexity of the problem.
Note that when degradations are added to $\mathbf{I}$, the visibility of the facial details may vary. Therefore, we adaptively replace the annotated mask with an all 1 mask (except for the hair and background regions) when the degradation is severe.

\noindent \textbf{Reference Selection Module.}
Given the $\{\mathbf{I}_\mathit{LQ}, \mathbf{I}^\mathit{Ref}, \mathbf{M}\}$ triplets, we can train our RefSel module, which is initialized with the U-Net in the data engine, \ie,
\begin{equation}
\mathbf{p} = f_\mathbf{M}([\mathbf{I}_\mathit{LQ},\mathbf{I}^\mathit{Ref}]; \Theta_\mathbf{M}),
\end{equation}
where $\mathbf{p}$ is then binarized to obtain the mask, and $f_\mathbf{M}(\cdot)$ and $\Theta_\mathbf{M}$ respectively denote the network structure and parameters of RefSel.
We employ the online hard example mining cross-entropy loss (OHEM-CE Loss) to measure the discrepancy between network-predicted masks and manually annotated ground truths, thereby driving the model to focus on challenging regions during the learning process. The loss function is defined by,
\begin{equation}
\mathcal{L}_{\text{OHEM-CE}} = -\tfrac{1}{|\mathbf{M}'|} {\Sigma}_{i \in \mathbf{M}'} m_i \log(p_i),
\end{equation}
where $\mathit{p}_\mathit{i}$ and $\mathit{m}_\mathit{i}$ are output and label for the position $\mathit{i}$, and $\mathbf{M}'$ denotes the regions whose loss is larger than a threshold $\tau$ (\ie, these regions are more difficult).

\begin{figure}[t]
	\centering
        \footnotesize
        \setlength\abovecaptionskip{0.2\baselineskip}
        \setlength\belowcaptionskip{0.2\baselineskip}
	\includegraphics[width=1.0\linewidth]{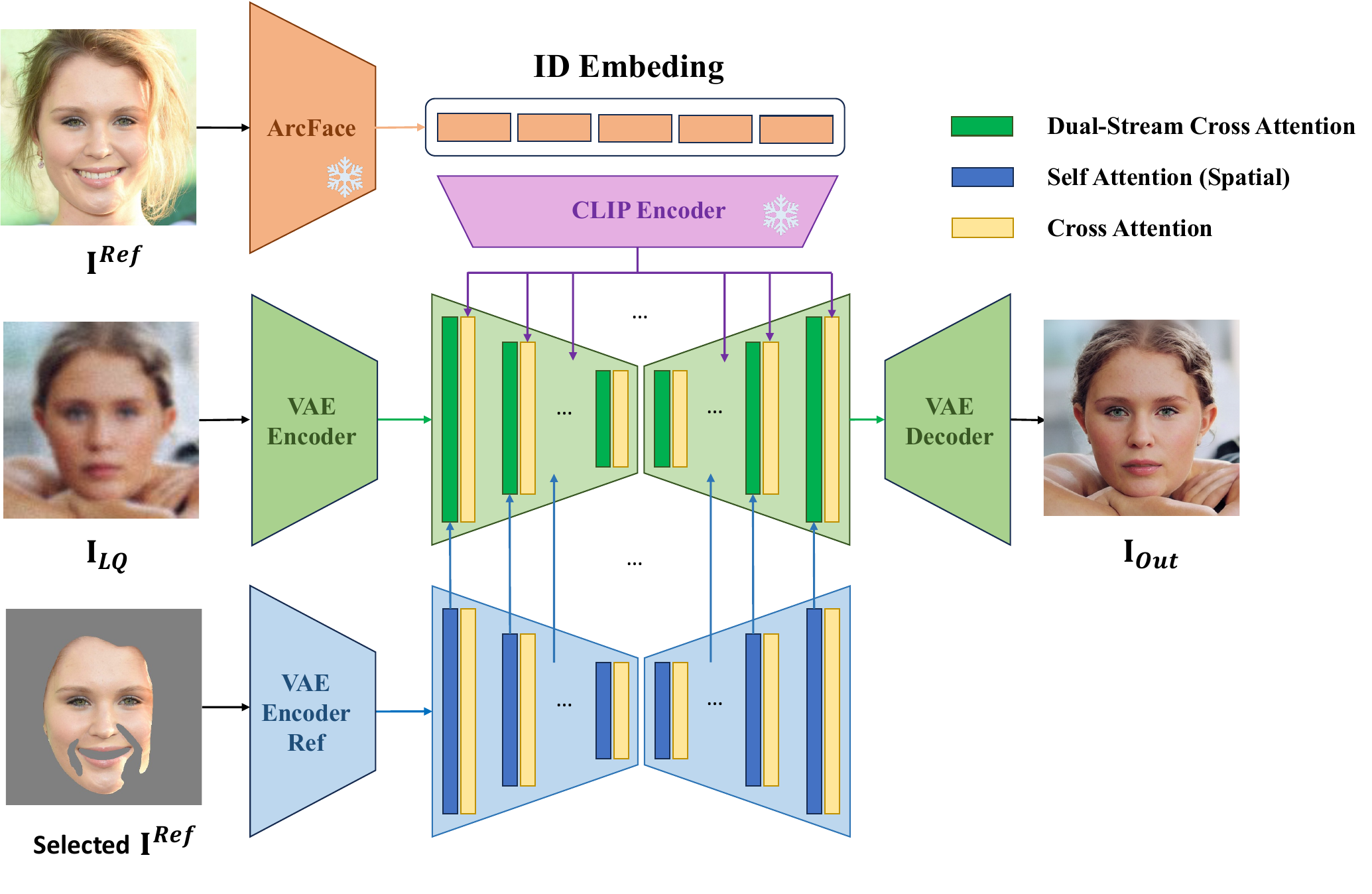}
    \caption{Overview of the network structure of our RefSTAR.}
    \label{fig:RefSTAR_restoration_network}
\end{figure}

\subsection{Reference Transfer in Restoration Model}
We build the restoration model upon a one-step diffusion framework based on Arc2Face~\cite{Arc2face}.
Note that the Arc2Face framework already introduced ArcFace embeddings fo the reference for identity preservation, which is insufficient to preserve detailed textures of the reference image.
Therefore, we consider referring to ReferenceNet~\cite{ReferenceNet} to better introduce reference features.
However, due to the similarity between different facial regions, the cross-attention deployed in ReferenceNet~\cite{ReferenceNet} cannot effectively match the input and reference features, resulting in a trivial solution in the cross-attention module.
Denote the features from the input by $Q_{in}$, $K_{in}$, and $V_{in}$, and features from the reference by $K_{ref}$ and $V_{ref}$, the vanilla cross-attnetion can be defined by,
\begin{equation}
    \sigma(\tfrac{Q_{in}[K_{in}, K_{ref}]^\top}{\sqrt{d_k}})[V_{in},V_{ref}].
    \label{eqn:RefSTAR_CA}
\end{equation}
It can be seen that when $Q_{in}$ and $K_{ref}$ are difficult to match, $Q_{in}K_{ref}^\top$ can be easily optimized towards $\mathbf{0}$, making the model ignore $V_{ref}$ in the final output.

\noindent\textbf{Dual-Stream Cross-Attention.}
To remedy the dilemma, we present a dual-stream cross-attention mechanism~\cite{IP-Adapter}.
Unlike \cref{eqn:RefSTAR_CA}, we process the input and reference features separately, then put them together to force the infusion, \ie,
\begin{equation}
    \sigma(\tfrac{Q_{in}K_{in}^\top}{\sqrt{d_k}})V_{in}+\sigma(\tfrac{Q_{in}K_{ref}^\top}{\sqrt{d_k}})V_{ref}.
    \label{eqn:RefSTAR_DSCA}
\end{equation}
In this way, the reference features are inevitably injected into the restoration model.

\begin{table*}[t]
    \centering
    \setlength\abovecaptionskip{0.25\baselineskip} 
    \setlength\belowcaptionskip{0.3\baselineskip}
    \setlength\tabcolsep{1mm}
    \caption{Quantitative comparison on synthetic and real world datasets. The methods are categorized into non-reference (w/o Reference) and reference-based (w/ Reference) face restoration for clarity. The best and second-best results are highlighted by {\color{red}\textbf{bold}} and {\color{blue}\underline{underline}}.}
    \small
    \setlength{\tabcolsep}{1.1mm}
	{
    \begin{tabular}{ccccccccccccc}
        \toprule
        & \multirow{2}{*}{Methods} & & \multicolumn{6}{c}{Celeb-Ref-Test} & \multicolumn{3}{c}{Real-World-Test} \\
        \cmidrule(r){4-9} \cmidrule(r){10-12}
        & & & PSNR$\uparrow$ & LPIPS$\downarrow$ & FID$\downarrow$ & ID-GT$\uparrow$ & ID-Ref$\uparrow$ & MUSIQ$\uparrow$ &  FID$\downarrow$ & ID-Ref $\uparrow$ & MUSIQ $\uparrow$ \\ 
        \midrule
        \multirow{5}{*}{w/o Reference} &  \multirow{3}{*}{GAN Based} 
        & GFPGAN & 24.56 & 0.367 & 24.01 & 76.60 & 44.02 & {\color{red}\textbf{74.55}} & 163.54 & 42.36 & {\color{blue}\underline{73.51}} \\ 
        & & GPEN & 24.31 & 0.414 & 27.95 & 79.25 & 45.54 & 74.02 & 162.55 & 44.58 & {\color{red}\textbf{74.29}} \\ 
        & & CodeFormer & 24.72 & {\color{blue}\underline{0.349}} & 24.63 & 74.49 & 43.33 & 74.35 & 159.30 & 38.39 & 73.03 \\
        \cmidrule{2-2}
        & \multirow{2}{*}{Diffusion Based}
        & OSEDiff & 23.46 & 0.354 & 37.01 & 62.77 & 36.82 & 74.32 & 167.84 & 37.15 & 73.42 \\ 
        & & DiffBIR & {\color{blue}\underline{24.93}} & 0.402 & 28.31 & {\color{blue}\underline{80.32}} & 46.37 & {\color{blue}\underline{74.44}} & 166.95 & 44.30 & 72.73 \\ 
        \midrule
        \multirow{5}{*}{w/ Reference} & \multirow{2}{*}{GAN Based} 
        & ASFFNet & 24.51 & 0.351 & 25.15 & 74.01 & 49.39 & 73.78 & 174.44 & 45.00 & 71.65\\ 
        & & DMDNet & {\color{red}\textbf{25.12}} & 0.374 & 33.49 & 77.09 & 48.59 & 71.97 & 163.44 & 45.34 & 72.04 \\
        \cmidrule{2-2}
        & \multirow{4}{*}{Diffusion Based}
        & FaceMe & 24.73 & 0.399 & 29.18 & 74.21 & 42.32 & 73.17 & 165.77 & 43.14 & 69.64 \\ 
        & & RefLDM & 24.16 & 0.387 & {\color{blue}\underline{22.90}} & 79.32 & {\color{blue}\underline{53.31}} & 73.03 & {\color{blue}\underline{155.73}} & {\color{blue}\underline{46.70}} & 71.10 \\ 
        & & RestoreID & 24.25 & 0.409 & 28.82 & 76.65 & 49.82 & 73.12 & 162.97 & 45.89 & 71.44 \\ 
        & & \textbf{Ours} & 24.69 & {\color{red}\textbf{0.335}} & {\color{red}\textbf{21.01}} & {\color{red}\textbf{82.76}} & {\color{red}\textbf{64.53}} & 73.91 & {\color{red}\textbf{155.18}} & {\color{red}\textbf{56.44}} & 72.75 \\ 
        \bottomrule
    \end{tabular}
    }
    \label{tab:RefSTAR_performance_comparison_all}
    \vspace{-6pt}
\end{table*}

\subsection{Mask-Compatible Cycle-Consistency Loss}
Considering that the restoration is finally supervised by the ground truth $\mathbf{I}$, solely a dual-stream cross-attention is less effective.
To encourage the reference features to appear in the final output, we deploy a cycle-consistency loss, which in turn takes the degraded reference as new input and the original output as new reference to reconstruct the original reference $\mathbf{I}^\mathit{Ref}$.
By applying the high-quality reference as the supervision, the detailed textures in the reference are required to consist in the original output.
Notably, with the binary mask, not all features of the reference are transferred to the final output. Therefore, the cycle-consistency loss is modulated to be mask-compatible, \ie,
\begin{equation}
    \mathcal{L}_\mathit{cycle}={\Sigma}_i\lambda_i\ell_i(\mathbf{M}\odot\mathbf{I}^{Ref}_{out}, \mathbf{M}\odot\mathbf{I}^{Ref}),
\end{equation}
where $\odot$ denoted entry-wise multiplication, $\ell_i$ denotes loss functions such as $\ell_1$ loss, identity loss, and perceptual loss. The other loss function alsocontains fidelity loss ($\ell_1$), perception loss~\cite{Real-ESRGAN}, ID loss~\cite{Arcface} and GAN loss~\cite{GAN}. The total loss is defined by,
\begin{equation}
\begin{split}
    \mathcal{L} = \lambda_1\mathcal{L}_1(\mathbf{I}_\mathit{Out},\mathbf{I}) + \lambda_2\mathcal{L}_\mathrm{perc}(\mathbf{I}_\mathit{Out},\mathbf{I}) +
    \lambda_3\mathcal{L}_\mathrm{id}(\mathbf{I}_\mathit{Out},\mathbf{I}) \\+ \lambda_4\mathcal{L}_\mathrm{adv}(\mathbf{I}_\mathit{Out}) + \lambda_5\mathcal{L}_\mathrm{cycle}(\mathbf{M}\odot\mathbf{I}^{Ref}_{out}, \mathbf{M}\odot\mathbf{I}^{Ref}).
    \label{eqn:RefSTAR_learning_objective}
\end{split}
\end{equation}

\section{Experiments}

\noindent\textbf{Training Data.}
For the restoration model, we follow these reference-based methods and use the CelebRef-HQ~\cite{DMDNet} dataset, which consists of 10,555 facial images at a resolution of 512$\times$512 across 1,005 identities. The degraded inputs are generated using the two-stage degradation pipeline of Real-ESRGAN~\cite{Real-ESRGAN}, with motion and defocus deblurring embedded for a more realistic degradation.
The reference images are randomly selected from each identity to enhance generalization on real-world scenarios.
For RefSel module, we apply the same degradation pipeline to convert ground-truth images into their corresponding degraded inputs.

\noindent\textbf{Testing Data.}
For evaluating the restoration model, we select 1,000 identities from the CelebA~\cite{CelebA} dataset, with one image randomly selected as the reference and another as the ground truth for each identity. 
We adopt the same degradation process as other methods~\cite{DiffBIR} to generate the testing data. To evaluate the performance in real-world scenarios, we collect 60 celebrity image pairs from the internet (\ie, RealRef60), each consisting of one high-resolution image as reference and one low-resolution image as degraded input.
These pairs are selected to ensure that there was no overlap with the CelebRef-HQ dataset.
To assess the performance of RefSel, we use 250 pairs from the RefSel-HQ dataset as the test set.

\noindent\textbf{Implementation Details.}
 The RefSel module employs the AdamW optimizer~\cite{adamw} with $\beta_1$ = 0.5 and $\beta_2$ = 0.999, and a learning rate of $1\times10^{-3}$. 
The restoration model is trained using a two-phase training strategy. In Phase I, we initialize the model pre-trained weights from Arc2Face~\cite{Arc2face} and adapt it into a one-step diffusion-based image restoration model by training a set of LoRA adapters. In Phase II, we freeze the LoRA parameters and train only the ReferenceNet and our DSCA module. 
Both phases use the AdamW optimizer with $\beta_1$ = 0.5 and $\beta_2$ = 0.999, with a learning rate of $1\times10^{-4}$.
All experiments were conducted on a server equipped with two NVIDIA A800 GPUs. 

\begin{figure*}
	\centering
        \footnotesize
        
        \begin{minipage}{0.94\linewidth}
		\centering
	   \includegraphics[width=1.0\linewidth]{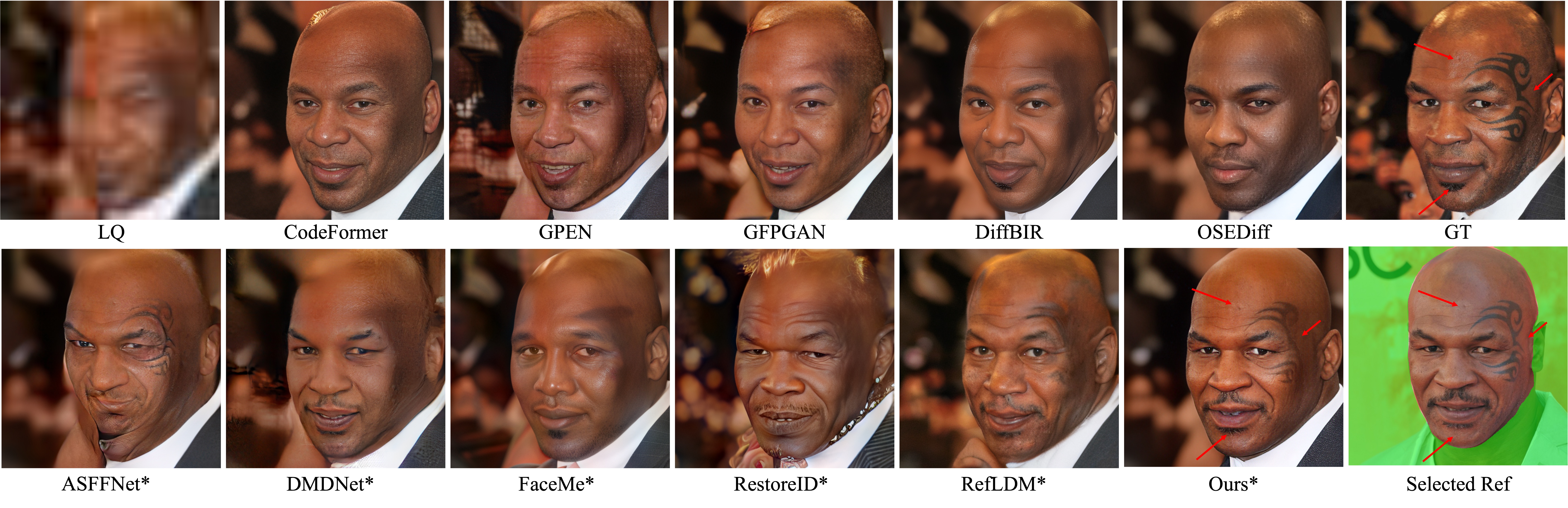}
        \end{minipage}
        \begin{minipage}{0.94\linewidth}
		\centering
	   \includegraphics[width=1.0\linewidth]{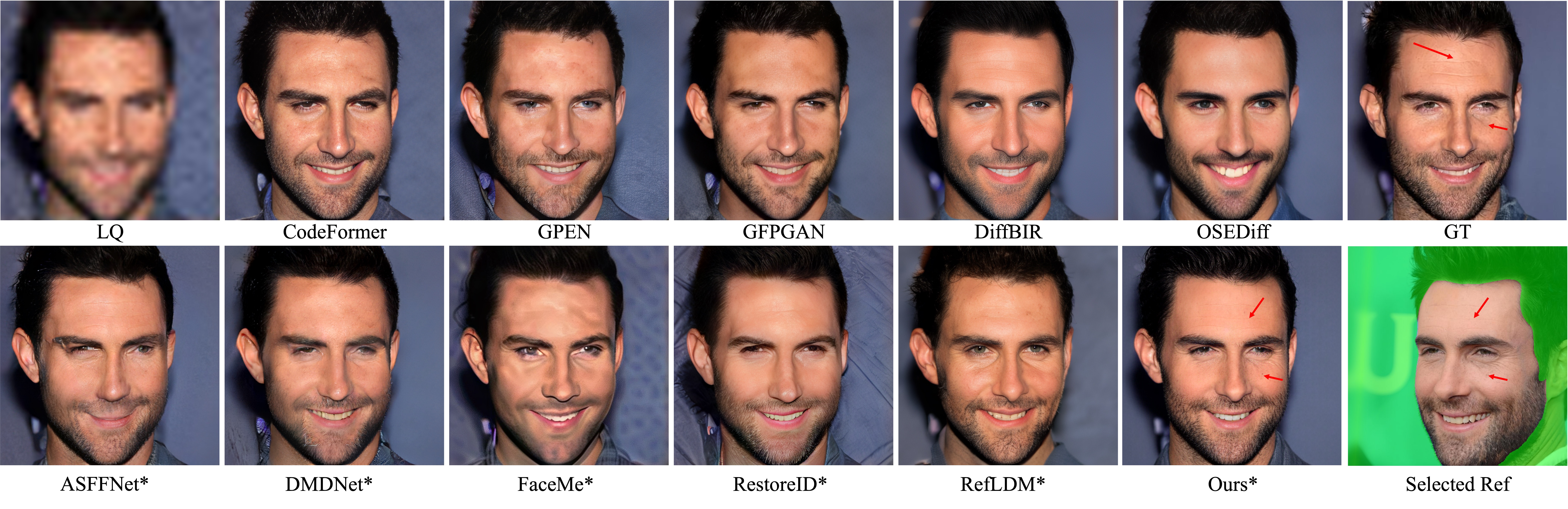}
        \end{minipage}
 
    \caption{Visual comparison against state-of-the-art blind face restoration methods and reference-guided face restoration methods on synthesized datasets. The methods with * denote the reference-guided face restoration methods.}
    \label{fig:RefSTAR_vis_syn}
    \vspace{-6pt}
\end{figure*}

\begin{figure*}
	\centering
        \footnotesize
        
        \begin{minipage}{0.94\linewidth}
		\centering
	   \includegraphics[width=1.0\linewidth]{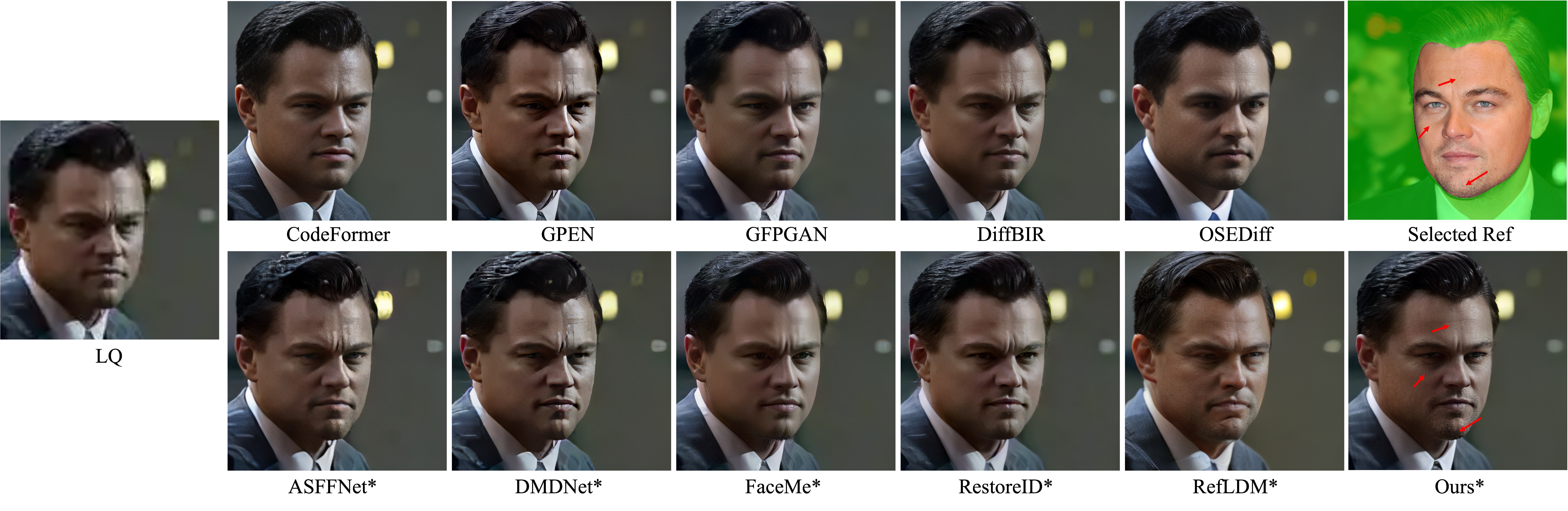}
        \end{minipage}
        \begin{minipage}{0.94\linewidth}
		\centering
	   \includegraphics[width=1.0\linewidth]{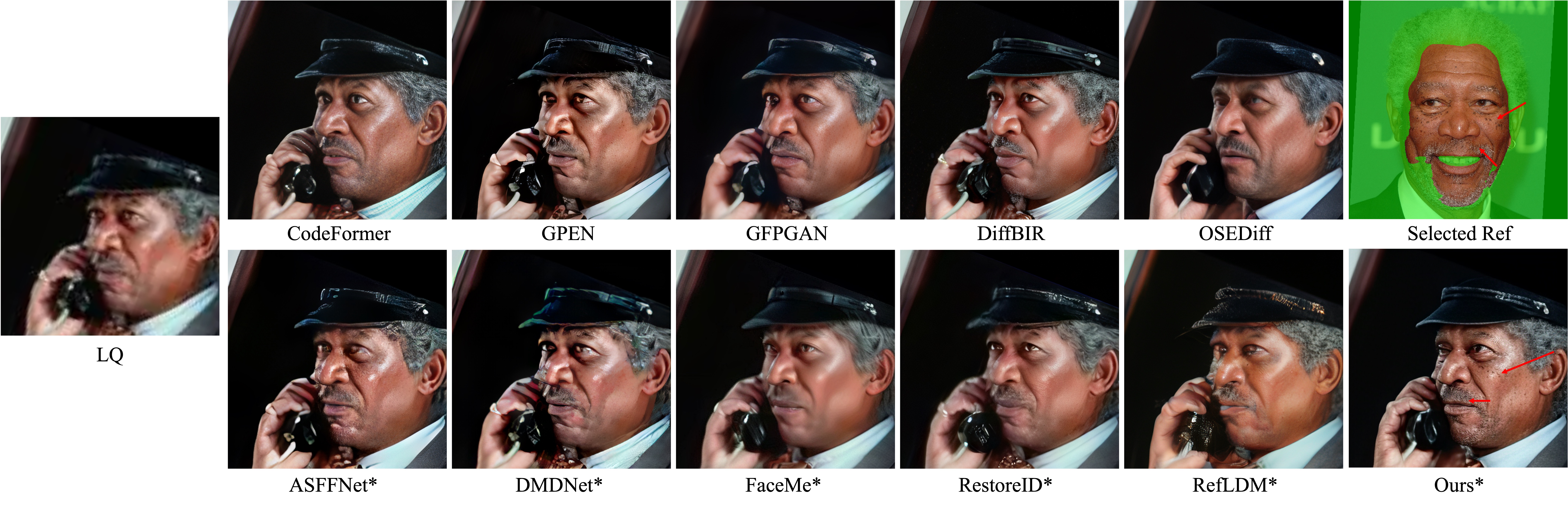}
        \end{minipage}
  
    \caption{Visual comparison against state-of-the-art blind face restoration methods and reference-guided face restoration methods on real world datasets. The methods with * denotes the reference-guided face restoration methods.}
    \label{fig:RefSTAR_vis_real}
\end{figure*}

\subsection{Comparison with Other Methods}

To demonstrate the superiority of our RefSTAR, we compare it with several representative blind face restoration methods, including non-reference methods (\ie, GPEN~\cite{GPEN}, GFP-GAN~\cite{GFPGAN},  CodeFormer~\cite{CodeFormer}, DiffBIR~\cite{DiffBIR}, and OSEDiff~\cite{OSEDiff})  and reference-based methods (\ie, ASFFNet~\cite{ASFFNet}, DMDNet~\cite{DMDNet},  RefLDM~\cite{RefLDM}, FaceMe~\cite{FaceMe}, and RestoreID~\cite{RestoreID}), which cover both GAN-based and Diffusion-based models.

\noindent\textbf{Quantitative Comparison.}
We evaluate the model using PSNR, LPIPS~\cite{LPIPS}, FID~\cite{FID}, and MUSIQ~\cite{MUSIQ}. To further assess its ability to preserve facial identity, we compute identity similarity scores using the ArcFace model~\cite{Arcface}.
We provide two distances: one between the restored images and the ground truth, and the other between the restored images and the reference images, denoted as ID-GT and ID-Ref, respectively.
Since ground-truth image is unavailable for the real-world dataset RealRef60,  we only use FID, ID-Ref, and MUSIQ as evaluation metrics. 
As for FID, it is calculated using ground-truth as reference set for synthetic datasets, while for real-world datasets, it uses 70,000 high-resolution images from FFHQ as the reference set.

The quantitative results on both synthetic and real-world test sets are summarized in~\cref{tab:RefSTAR_performance_comparison_all}. 
For the synthetic testset, our approach achieves superior performance across most metrics, including LPIPS, FID, ID-GT, and ID-Ref. 
Our method performs on par with existing methods in PSNR and MUSIQ, while significantly outperforming them in LPIPS and FID, demonstrating superior visual quality in restored images.
Notably, our method achieves the highest ID-GT and ID-Ref scores among all reference-based blind face restoration approaches, demonstrating the effect of our method in transferring reliable reference features, leading to superior identity preservation.
For the real-world dataset, our method consistently outperforms others in FID and ID-Ref, while achieving comparable MUSIQ scores. This demonstrates its effectiveness in handling complex degradations and highlights its great generalization capability across different real-world scenarios.

To evaluate the performance of RefSel in identifying texture-consistent regions, we categorize the analysis into five distinct scenarios: (1) mouth open/close, (2) eyes open/close, (3) wrinkles conflicts, (4) beard conflicts, and (5) conflicts with glasses and patterns. 
Each scenario consists of 50 paired images. 
The evaluation results for RefSel are shown in~\cref{tab:RefSTAR_refsel_eval}. The accuracy demonstrates the reliability and effectiveness of our RefSel module, showcasing its great ability to distinguish regions with consistent texture and effectively identify conflicts between the input and reference.

\begin{table}[t]
    \centering
    \small
    \setlength\abovecaptionskip{0.2\baselineskip} 
    \setlength\belowcaptionskip{0.3\baselineskip}
    \setlength\tabcolsep{1mm}
    \renewcommand\arraystretch{1.1}
    \caption{The accuracy of our RefSel module on five scenarios: (1)~mouth open/close, (2) eyes open/close, (3) conflict wrinkles, (4)~conflict beard, and (5) conflict glasses and patterns.}
    \label{tab:RefSTAR_refsel_eval}
    \small
    \setlength{\tabcolsep}{1.1mm}
	{
        \begin{tabular}{ccccccc}
        \toprule
         & (1) & (2) & (3) & (4) & (5) & Average \\
        \midrule
        Accuracy & 0.90 & 0.90 & 0.86 & 0.94 & 0.82 & 0.88  \\
         \bottomrule
        \end{tabular}}
    \vspace{-3mm}
\end{table}

\noindent\textbf{Qualitative Comparison.}
Figs.~\ref{fig:RefSTAR_vis_syn} and \ref{fig:RefSTAR_vis_real} show the restoration results on synthetic and real-world degraded images, respectively. We can find that our RefSTAR consistently demonstrates more faithful restoration performance. Specifically, for challenging personalized textures such as tattoos, RefSTAR excels in transferring features from the reference image. The framework also accurately replicates common skin textures, including wrinkles and age spots, with this precise texture transfer being crucial for superior identity retention. Notably, our model maintains robust texture transfer even under significant pose variations between the source and reference images, demonstrating the effectiveness of our RefSTAR in unconstrained scenarios.

\begin{figure}[t]
	\centering
        \footnotesize
       
	\includegraphics[width=1.0\linewidth]{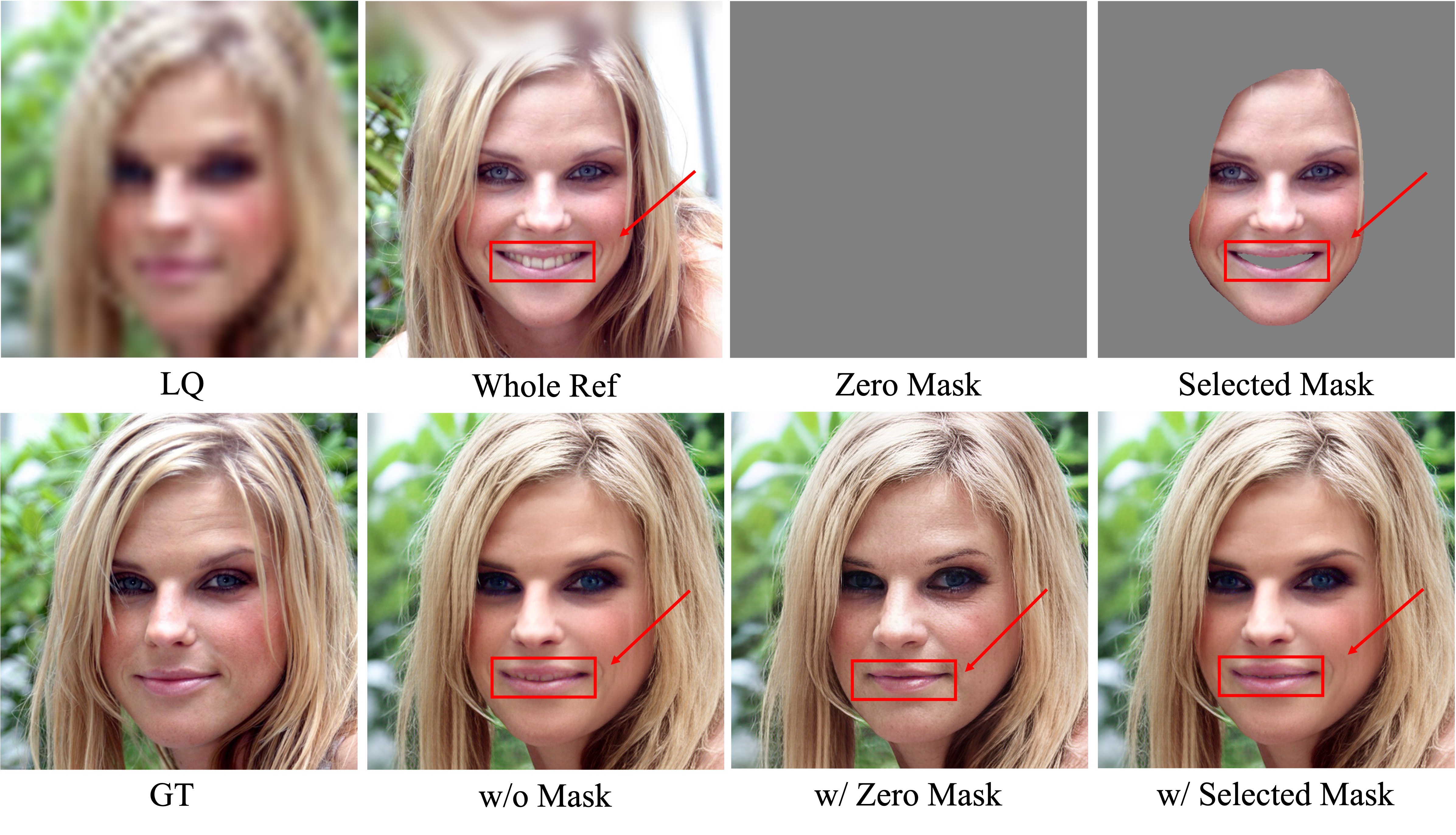}
    \caption{The visual comparison w/ or w/O RefSel.}
    \label{fig:RefSTAR_vis_ablation_refsel}
    \vspace{-5mm}
\end{figure}

\subsection{Ablation Studies}

\noindent\textbf{The Effect of RefSel.}
We analyze the effect by applying different binary masks (all-zeros or all-ones) to reference images.
As shown in~\cref{fig:RefSTAR_vis_ablation_refsel}, there is an inconsistency in the mouth region (open/close) due to different expressions.
With an all-ones mask (using the whole reference image), the unmasked teeth area in the reference image is improperly integrated into the output.
When all reference image is ignored by using an all-zeros mask for ReferenceNet, the output relies solely on the highly compressed ID embedding from ArcFace, leading to a loss of personalized textures and a slight decline in identity preservation.
The accurate selection by RefSel ensures that only consistent reference features are incorporated into the restored result, preserving wrinkles while masking the teeth. \cref{fig:RefSTAR_refsel_vis} shows more prediction results on different scenarios.

\cref{tab:RefSTAR_ablation_refsel} further demonstrates that the model with the all-ones mask achieves the highest ID-Ref score, as it incorporates the most information from the reference. However, the model with the selected mask achieves the highest ID-GT, demonstrating that our RefSel effectively selects the features most relevant to the input, resulting in more reasonable personalized texture transfer.

\begin{table}[t]
    \centering
    \small
    \setlength\abovecaptionskip{0.2\baselineskip} 
    \setlength\belowcaptionskip{0.3\baselineskip}
    \setlength\tabcolsep{1mm}
    \renewcommand\arraystretch{1.2}
    \caption{Ablation study on reference selection (RefSel).}
    \label{tab:RefSTAR_ablation_refsel}
    \scalebox{0.88}{
        \begin{tabular}{ccccccc}
        \toprule
        Methods & PSNR$\uparrow$ & LPIPS$\downarrow$ & FID$\downarrow$ & ID-GT$\uparrow$ & ID-Ref$\uparrow$ & MUSIQ$\uparrow$\\
        \midrule
        w/ all-ones mask & 24.65 & 0.342 & 21.52 & 81.86 & {\color{red}\textbf{66.61}} & 73.88 \\
        w/ all-zeros mask & 23.89 & 0.343 & 23.13 & 77.70 & 44.39 & {\color{red}\textbf{74.73}} \\
        w/ seleted mask & {\color{red}\textbf{24.69}} & {\color{red}\textbf{0.335}} & {\color{red}\textbf{21.01}} & {\color{red}\textbf{82.76}} & 64.53 & 73.91 \\
         \bottomrule
        \end{tabular}}
    \vspace{-1mm}
\end{table}

\begin{figure}[h]
	\centering
        \footnotesize
        \setlength\abovecaptionskip{2pt}
        \setlength\belowcaptionskip{0pt}
	\includegraphics[width=.9\linewidth]{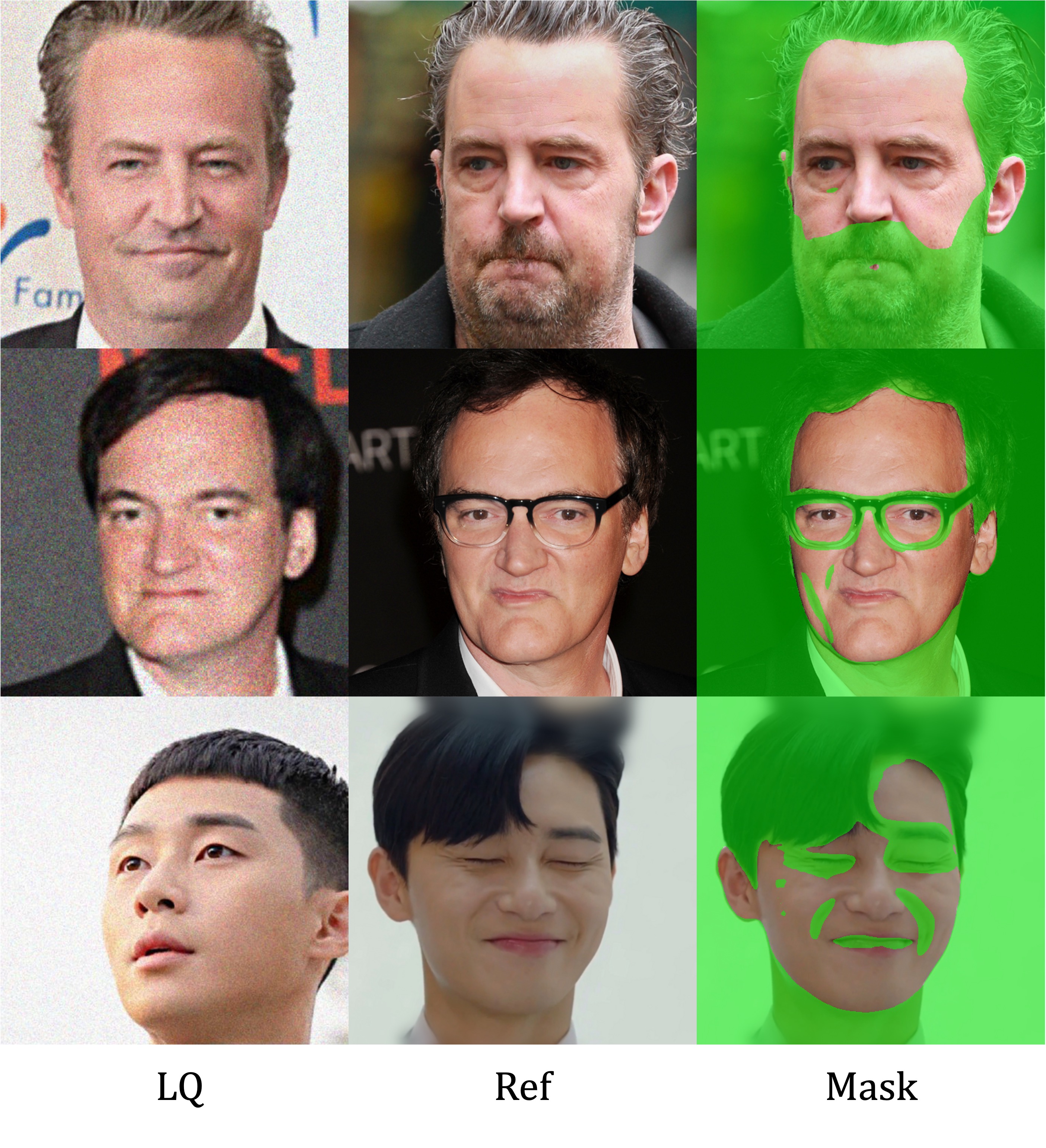}
    \caption{The visualization of predicted texture-consistent regions.}
    \label{fig:RefSTAR_refsel_vis}
\end{figure}

\begin{figure}
	\centering
        \footnotesize
        \setlength\abovecaptionskip{5pt}
        \setlength\belowcaptionskip{3pt}
	\includegraphics[width=.9\linewidth]{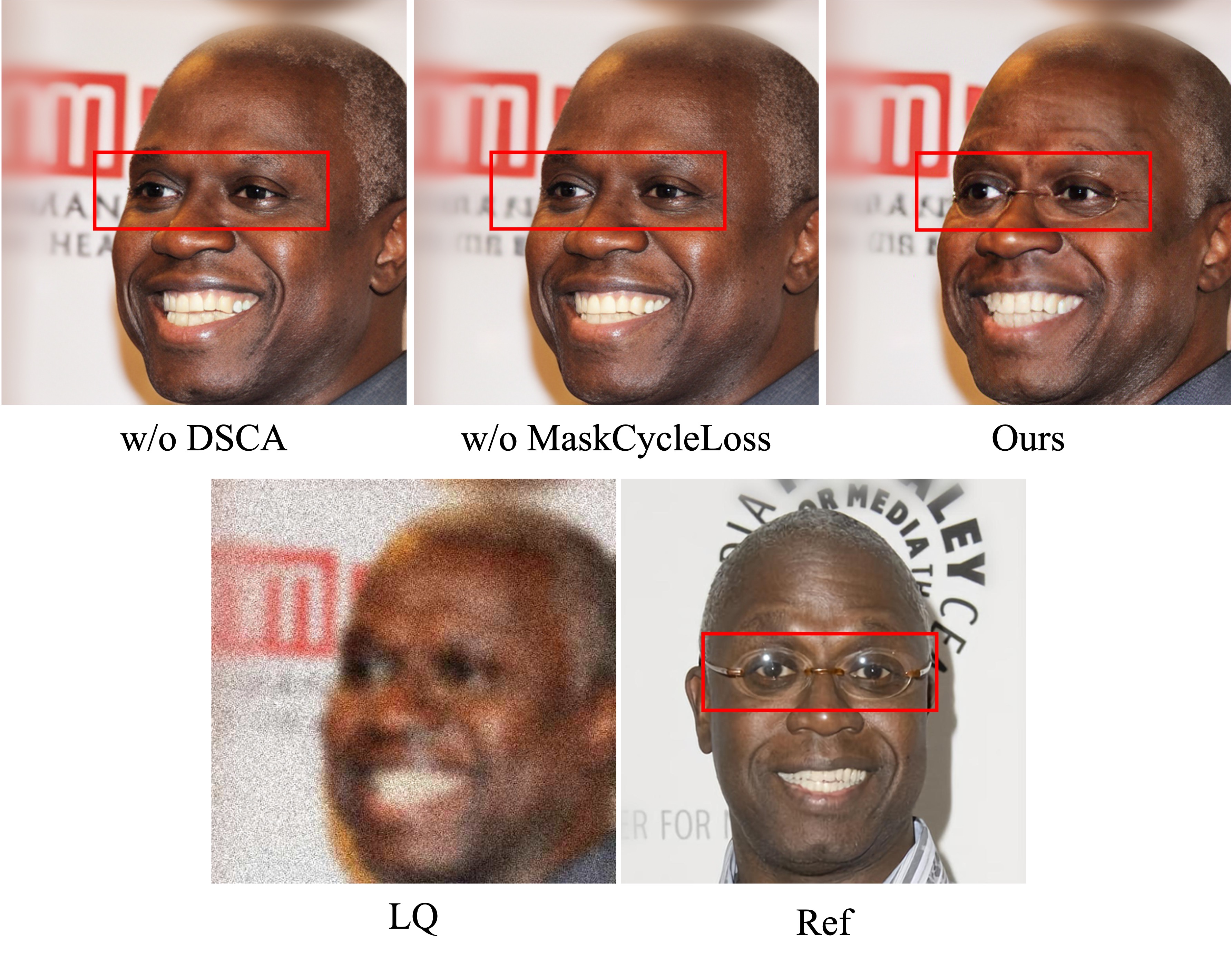}
    \caption{Comparison w/ or w/o DSCA and Cycle-Consistency Loss.}
    \label{fig:RefSTAR_vis_ablation_cycle}
\end{figure}

\noindent\textbf{The Effect of Dual-Stream Cross-Attention.}
Instead of using the cross-attention module with concatenated $K$ and $V$, we propose a dual-stream attention mechanism. \cref{fig:RefSTAR_vis_ablation_cycle} shows that the previous attention mechanism struggles to effectively incorporate reference features into the output, even when Mask-Compatible Cycle-Consistency Loss is applied. This issue arises because image detail restoration requires stricter alignment between the source and reference inputs. As a result, the network behavior becomes biased, and the concatenation of the $K$ and $V$ of the input and reference causes the network to ignore the reference, even with the softmax function. \cref{tab:RefSTAR_ablation_cycle_loss} also indicates that it is challenging to properly leverage reference textures without DSCA module.

\noindent\textbf{The Effect of Cycle-Consistency Loss.}
To improve the integration of textural details from reference images, we propose a cycle-consistency loss with a masking mechanism during training. To evaluate the effect, we conduct an ablation study by removing it from the pipeline.
As shown in~\cref{fig:RefSTAR_vis_ablation_cycle}, incorporating cycle-consistency loss significantly enhances the model to capture textures from reference images, while also improving identity-aware feature alignment. 
The cycle-consistency loss constraint ensures that the restored image more faithfully aligns with the reference, enabling better texture transfer and more accurate image restoration. \cref{tab:RefSTAR_ablation_cycle_loss} further validates that this loss function significantly enhances the performance of reference utilization. By ensuring the reconstruction aligns more closely with the reference image, the cycle-consistency loss not only improves texture transfer but also strengthens identity preservation. This results in more accurate feature alignment, effectively bridging the gap between the input and reference and thus boosting overall restoration quality.

\begin{table}[t]
    \vspace{-4pt}
    \centering
    \small
    \setlength\abovecaptionskip{0.2\baselineskip} 
    \setlength\belowcaptionskip{0.3\baselineskip}
    \setlength\tabcolsep{1.2mm}
    \renewcommand\arraystretch{1.2}
    \caption{Ablation study on DSCA and Cycle-Consistency Loss.}
    \label{tab:RefSTAR_ablation_cycle_loss}
    \scalebox{0.88}{
        \begin{tabular}{ccccccc}
        \toprule
        Methods & PSNR$\uparrow$ & LPIPS$\downarrow$ & FID$\downarrow$ & ID-GT$\uparrow$ & ID-Ref$\uparrow$ & MUSIQ$\uparrow$\\
        \midrule
        Ours & 24.69 & {\color{red}\textbf{0.335}} & {\color{red}\textbf{21.01}} & {\color{red}\textbf{82.76}} & {\color{red}\textbf{64.53}} & 73.91 \\
        w/o cycle & {\color{red}\textbf{25.10}} & 0.341 & 22.66 & 80.01 & 60.70 & 74.45 \\
        w/o DSCA & 23.73 & 0.350 & 27.84 & 74.78 & 60.33 & {\color{red}\textbf{74.97}} \\
         \bottomrule
        \end{tabular}}
    \vspace{-1mm}
\end{table}

\noindent\textbf{The Effect of Using Pre-Trained Arc2Face.}
To enhance identity priors, our base model incorporates the pre-trained Arc2Face architecture, which is integrated into a one-step diffusion framework using a set of LoRA. To assess the performance of the Arc2Face prior in preserving facial identity from reference images, we conduct a comparative experiment by replacing Arc2Face with Stable Diffusion v1.5 (SD1.5) as the base model.
As shown in Tab.~\ref{tab:RefSTAR_ablation_face_model}, evaluations on the CelebA-Ref-Test dataset demonstrate that the Arc2Face-based model significantly outperforms the SD1.5 variant in both ID-GT and ID-Ref. This underscores the importance of the identity-aware priors embedded in Arc2Face for effectively integrating facial identity cues from reference images into the restoration process.

\begin{table}[t]
    \centering
    \small
    \setlength\abovecaptionskip{0.2\baselineskip} 
    \setlength\belowcaptionskip{0.3\baselineskip}
    \setlength\tabcolsep{1mm}
    \renewcommand\arraystretch{1.2}
    \caption{Ablation study on the base model. }
    \label{tab:RefSTAR_ablation_face_model}
    \scalebox{0.88}{
        \begin{tabular}{ccccccc}
        \toprule
        Methods & PSNR$\uparrow$ & LPIPS$\downarrow$ & FID$\downarrow$ & ID-GT$\uparrow$ & ID-Ref$\uparrow$ & MUSIQ$\uparrow$\\
         \midrule
         w/ Arc2Face & {\color{red}\textbf{24.69}} & {\color{red}\textbf{0.335}} & {\color{red}\textbf{21.01}} & {\color{red}\textbf{82.76}} & {\color{red}\textbf{64.53}} & {\color{red}\textbf{73.91}} \\
        w/ SD1.5 & 23.71 & 0.365 & 25.90 & 72.38 & 62.61 & 73.83 \\
        
         \bottomrule
    \end{tabular}}
    \vspace{-2mm}
\end{table}

\section{Conclusion}
\label{sec:conclusion}
In this work, we introduce RefSTAR, a reference-based blind facial image restoration framework that effectively addresses reference selection, transfer, and reconstruction. Our RefSel module ensures accurate texture consistency selection, while the structured feature fusion mechanism boosts the reference transfer. Additionally, the cycle consistency constraint further enhances identity preservation and restoration fidelity.
By incorporating manually annotated consistency regions, our RefSel-HQ dataset emphasizes the critical role of supervised guidance in enhancing reference-based restoration.
Experiments demonstrate that RefSTAR outperforms existing methods in maintaining identity-related features and generating high-fidelity restorations. 

{
    \small
    \bibliographystyle{ieeenat_fullname}
    \bibliography{main}
}

\end{document}